\documentclass{sn-jnl}
\usepackage[super]{natbib}
\usepackage[utf8]{inputenc}
\UseRawInputEncoding
\usepackage{graphicx}
\usepackage{booktabs}
\usepackage{longtable}
\usepackage{multirow}
\usepackage{makecell}
\usepackage{subcaption}
\usepackage{colortbl}
\usepackage{float}

\usepackage{amsmath,amssymb,amsfonts}
\usepackage{amsthm}
\usepackage{mathrsfs}

\usepackage[title]{appendix}

\usepackage{algorithm}
\usepackage{algorithmicx}
\usepackage{algpseudocode}
\usepackage{listings}

\usepackage{textcomp}
\usepackage{manyfoot}
\usepackage{pifont}
\usepackage{verbatim}
\usepackage{mdframed}
\usepackage{gensymb}
\usepackage{comment}
\usepackage{nopageno}
\usepackage{lineno}

\usepackage{xcolor}
\usepackage{hyperref}
\hypersetup{
    colorlinks=true,
    linkcolor=blue,
    filecolor=magenta,
    urlcolor=cyan,
    pdftitle={Causality4FoodSecurity},
    pdfpagemode=FullScreen,
}

\definecolor{nice-green}{HTML}{007849}
\definecolor{nice-blue}{HTML}{0375B4}
\definecolor{nice-orange}{HTML}{CC7722}
\definecolor{nice-red}{HTML}{FF5733}
\definecolor{mutedpurple}{HTML}{9467BD}

\definecolor{cblue}{RGB}{164,220,255}
\definecolor{corange}{RGB}{255,187,135}
\definecolor{cgreen}{RGB}{0,139,0}
\definecolor{psred}{RGB}{246,159,159}
\definecolor{psblue}{RGB}{142,195,255}

\newcommand\setrow[1]{\gdef\rowmac{#1}#1\ignorespaces}
\newcommand\clearrow{\global\let\rowmac\relax}

\begin{document}
%\linenumbers

\UseRawInputEncoding

\title[Credit Access Improves Food Security]{
Credit Access Is Associated with Improved Food Security in the Horn of Africa
}

\author*[1]{Jordi Cerd\`a-Bautista}\email{jordi.cerda@uv.es}
\author[1,2]{Vasileios Sitokonstantinou}
\author[3]{Jos\'e Manuel Veiga L\'opez-Pe\~na}
\author[4]{Duccio Piovani}
\author[1,5]{Jos\'e Mar\'ia T\'arraga}
\author[1]{Gustau Camps-Valls}

\affil[1]{\orgdiv{Image Processing Laboratory}, \orgname{Universitat de Val\`encia},\country{Spain}}
\affil[2]{\orgdiv{Artificial Intelligence Group}, \orgname{Wageningen University \& Research}, \city{Wageningen}, \country{The Netherlands}}
\affil[3]{\orgdiv{Food Security Unit}, \orgname{Joint Research Centre (JRC), European Commission}, \city{Ispra}, \country{Italy}}
\affil[4]{\orgdiv{Early Warning \& Forecasting Unit}, \orgname{World Food Programme (WFP), United Nations}, \city{Rome}, \country{Italy}}
\affil[5]{\orgname{
Internal Displacement Monitoring Centre (IDMC)}, \city{Geneva}, \country{Switzerland}}

\abstract{
The intensification of climate change poses a growing threat to food security, especially in vulnerable communities.
This study employs an observational machine-learning framework to estimate the causal association between access to credit and acute food insecurity in Somalia and across the Horn of Africa, drawing on a harmonized dataset spanning key environmental, socioeconomic, and conflict-related factors from 2015 to 2022. Results indicate that greater credit access is associated with a 2\% reduction in acute food insecurity at the population level over the study period. Given that, on average, 16\% of the population is in crisis, this effect represents a meaningful shift within the at-risk group. We interpret these estimates under explicit identification assumptions and complement them with robustness and refutation tests. The results provide context-specific evidence on how financial access correlates with food security outcomes in data-scarce, crisis-affected settings, and offer a transparent framework for integrating heterogeneous data sources when randomized evaluations are infeasible.
}

\begin{nolinenumbers}
\maketitle
\end{nolinenumbers}
\clearpage

%\tableofcontents

\clearpage

%=======================================================
% 1. INTRODUCTION
%=======================================================

\begin{nolinenumbers}
\section*{Introduction}
\end{nolinenumbers}

The accelerating pace of climate change has intensified the frequency and severity of droughts, posing unprecedented challenges to food security in vulnerable regions worldwide \citep{kroeger2023heat, dasgupta2022attributing}. Communities reliant solely on rainfall for food production are increasingly at risk, often necessitating immediate humanitarian assistance to survive \citep{Funk2018, Pape2019}. Failure to act promptly can lead to severe economic losses, large-scale displacement, infant malnutrition, and heightened mortality rates due to hunger and famine \citep{Desai2021, SomaliaDisplaced2022, Maxwell2011}. Meanwhile, humanitarian organizations struggle with a widening gap between available funding and the growing needs of affected populations \citep{WFP2023, FAO2023, kagin2024cost}. Designing effective interventions under resource constraints is thus imperative, yet rigorous evidence identifying the most cost-effective strategies to assist populations in emergencies remains scarce \citep{Shannon2017}.

East Africa, particularly Somalia, has witnessed a concerning surge in acute food insecurity between 2022 and 2023, affecting at least 64 million people \citep{WFP2023, GRFC2024}. While prolonged droughts significantly contribute to this crisis \citep{Coughlan2019}, they are not the sole drivers. Factors such as hydrological conditions, agricultural capabilities, market access, insufficient aid, conflicts, and displacement also play critical roles \citep{Maxwell2023, Sneyers2017, Warsame2023, Abel2019,grijalva2018, Ronco2023, tarraga2024}. Understanding food security in this context is complex, involving multiple variables, spatial and temporal scales, and non-linear relationships. In crisis-affected regions, credit interventions or cash-based transfers are increasingly used as flexible, scalable forms of humanitarian assistance. Still, their real-world effectiveness remains difficult to quantify due to the complexity of overlapping environmental, social, and economic drivers. Traditional machine learning (ML) is widely used to monitor Earth and socio-environmental systems because it enables efficient processing of heterogeneous observational datasets. This enables us to map current ecological conditions and anticipate changes. However, these correlation-based models are not designed to address interventional and counterfactual questions \citep{pearl_2009, Peters2017, Runge23causalreview}, which are crucial for policy-making and humanitarian assistance.
This study employs observational causal inference methods to estimate the relationship between access to credit and acute food insecurity during severe droughts in the Horn of Africa.

Importantly, these methods are designed for observational settings and can be applied in other low-data, high-uncertainty environments where randomized trials are infeasible \citep{pearl_2009, Peters2017, Runge23causalreview, sharma2020dowhy}.
By identifying causal links between financial support and food outcomes, we contribute not just to understanding a single regional crisis but also to developing tools to improve resilience in food systems globally.

For this, we collected and harmonized a unique and extensive dataset of many societal, economic, environmental, and conflict factors in the region during the last decade. 
By estimating the effect of access to credit on acute food insecurity using causal machine learning, we provide robust evidence to enhance the effectiveness of humanitarian interventions and promote transparency in aid distribution.

\begin{figure}[t!]
\centerline{\includegraphics[width=12cm]{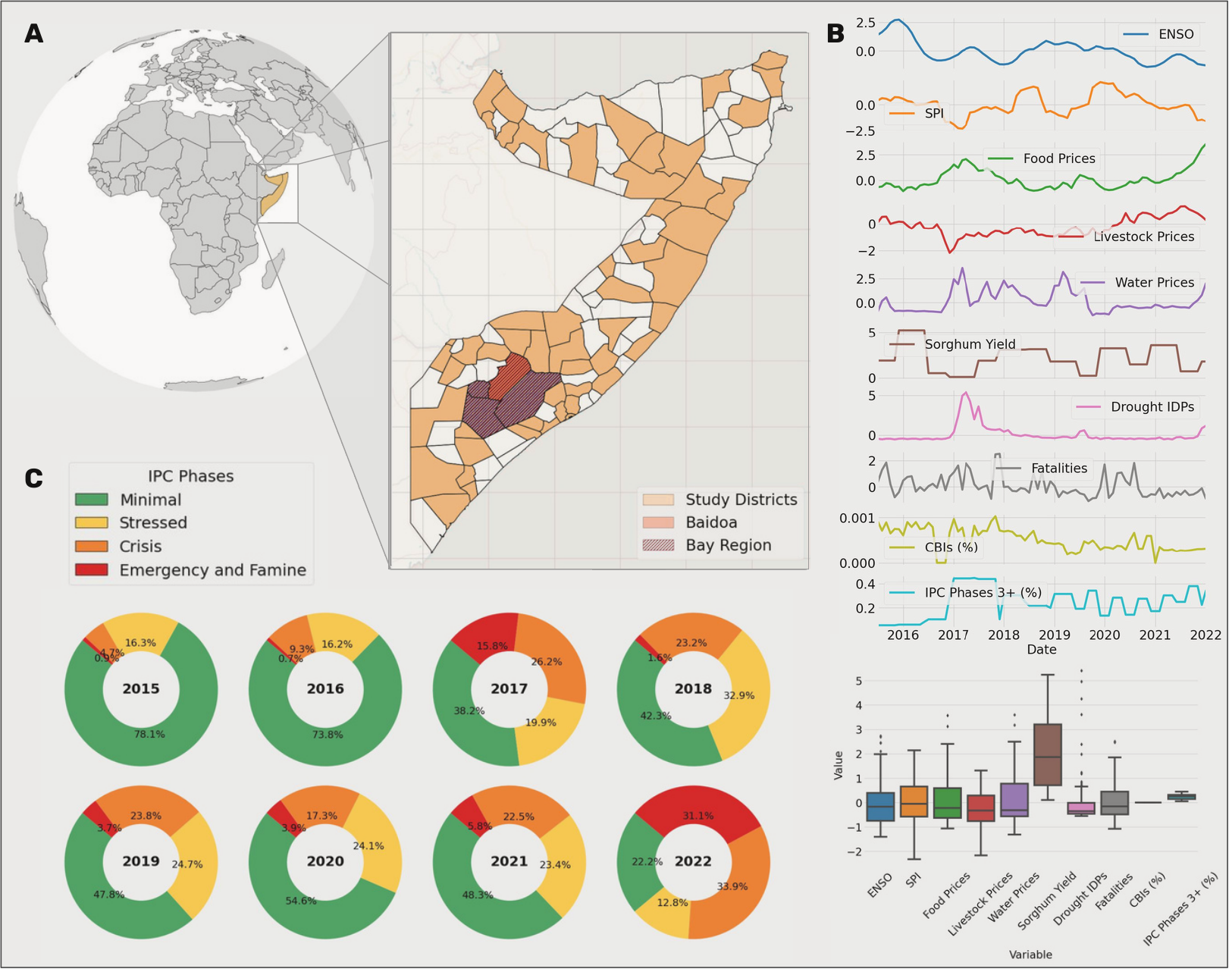}}
\caption{\textbf{Study area and data sources.} 
(A) Map of Somalia showing districts with available data, highlighting Baidoa, an important city in the Bay area (shown in the map), as an example. 
(B) Standardized time series and box plots displaying key environmental and socioeconomic variables for Baidoa, illustrating temporal trends and variability. 
(C) The annual distribution of IPC phases (percentages) for Baidoa shows shifts in food security over time.}

\label{fig:data}
\end{figure}

Despite the growing preference for cash-based interventions in humanitarian crises \citep{Shannon2017, Thomas2024} and the substantial literature evaluating the impacts of cash transfers \citep{attanasio2014using, haushofer2018long, mcintosh2024cash}, credit interventions \citep{crosta2024unconditional, nakano2020impact}, and other financial service interventions \citep{blumenstock2015predicting, suri2016long}, rigorous evaluations remain relatively scarce in fragile, data-sparse, and conflict-affected settings such as the Horn of Africa, where traditional experimental designs are often infeasible \citep{LENTZ2013151}. Previous studies have shown mixed results regarding their impact on food security  \citep{grijalva2018, burchi2018, hoddinott2018, McCullough2024}, and the need for geographically targeted interventions \citep{Ayalew2024}. This gap underscores the need for robust analytical approaches that can navigate the complexities of socio-environmental systems and provide actionable insights. By leveraging diverse data sources (including climatic, Earth observation, and socioeconomic datasets), our study employs modern observational causal inference methods, primarily the ML-based $X$-Learner, to evaluate the effectiveness of access to credit in mitigating acute food insecurity in Somalia (see \textit{Methods} section, cf. Fig.~\ref{fig:data}). Importantly, by relying on causal effect estimation methods, our analysis can assess the impact of a treatment variable and the complex interactions between the involved and often mediating processes and drivers of food security \citep{Runge23causalreview}.

While this study focuses on Somalia, the methodology is broadly applicable as an observational framework for integrating heterogeneous climate and socioeconomic data to evaluate interventions. Because our approach relies on observational data, it provides a framework for evaluating humanitarian interventions in other parts of the world where such data are available. However, the external validity of estimated effects remains context-dependent, and any policy interpretation should be made with caution, given measurement and identification limitations. \vspace{0.5cm}

%=======================================================
% 2. METHODS
%=======================================================

\begin{nolinenumbers}
\section*{Methods}
\label{sec:methods}
\end{nolinenumbers}

\begin{nolinenumbers}
\subsection*{\em Data collection and setup}
\end{nolinenumbers}

We collect and harmonize observational data from multiple sources. The outcome variable is the Integrated Food Security Phase Classification (IPC) \citep{ipc_manual2021}, defined as {\em ``food deprivation that threatens lives or livelihoods, regardless of the causes, context or duration''}. The IPC classifies the severity of food insecurity across five phases, from minimal acute food insecurity to famine. We focus on Phase 3 (Crisis) or worse (IPC 3+), which requires urgent humanitarian intervention, in line with the World Food Summit definition of food security as a multivariate, dynamic, and complex socio-environmental system \cite{WFS2009}. In Somalia, IPC data at the district level is provided by the Food Security and Nutrition Analysis Unit \citep{FSNAU}.
IPC is the most widely adopted food insecurity framework for Somalia, offering standardized, policy-relevant classifications used directly by humanitarian organizations. However, it is not a purely observational dataset: classifications are derived through expert consensus that combines quantitative indicators with qualitative assessments, and their construction varies across districts and over time due to differences in data availability and field access. This introduces potential spatial and temporal heterogeneity into the outcome variable \citep{Lentz2019399, Zhou2021, Lentz2024IPCAccuracy}.

For credit access, we use a proxy variable that represents the number of individuals receiving credit \citep{FSNAU}. Even though these data are among the most complete available for the Somali context and were selected for their policy relevance and coverage, we acknowledge that they may introduce bias if unmeasured or collinear factors influence both treatment and outcome.

Additional covariates include: ENSO phases from the World Meteorological Organization based on the Oceanic Niño Index \citep{Sazib}; a Standardized Precipitation Index built from CHIRPS \citep{SPI2012}; monthly market prices for livestock, staple food, and water, as well as sorghum yield estimates from FSNAU \citep{FSNAU}; conflict fatalities from ACLED \citep{ACLED}; remittances from FSNAU; and drought-induced internal displacement from UNHCR PRMN \citep{PRMN}. Data cover 56 Somali districts from July 2015 to January 2022 (see Table~\ref{tab:data} and Fig.~\ref{fig:data}). The 16 excluded districts had incomplete reporting; a comparison of IPC levels and population distributions revealed no significant systematic differences relative to the included districts (see \textit{Appendix A}).

\begin{table}[t]
\centering
\small
\caption{{\bf Variables and sources used in the study, spatial and temporal resolutions.}}
\begin{tabular}{llll}
\hline
Variable & Source & Spatial Resolution & Temp. Resolution \\
\hline
\rowcolor[HTML]{EFEFEF} 
ENSO & \href{WMO}{https://climexp.knmi.nl/selectdailyindex.cgi?id=someone@somewhere} & Country & Daily \\
SPI & \href{CHIRPS}{https://developers.google.com/earth-engine/datasets/catalog/UCSB-CHG_CHIRPS_DAILY#bands} & $0.05\degree$ & Daily \\
\rowcolor[HTML]{EFEFEF} 
Violent Conflict & \href{ACLED}{https://acleddata.com/#/dashboard} & Geolocated Event & Hourly\\
Local Market Prices & \href{FSNAU}{https://fsnau.org/ids/dashboard.php} & District & Monthly \\
\rowcolor[HTML]{EFEFEF} 
Sorghum Production & \href{FSNAU}{https://crops.fsnau.org/} & District & Seasonal \\
Drought Displacement & \href{UNHCR}{https://unhcr.github.io/dataviz-somalia-prmn/index.html} & District & Weekly\\
\rowcolor[HTML]{EFEFEF} 
Somalia Districts & \href{UNDP}{https://data.humdata.org/dataset/cod-ab-som} & District &  Static\\
Population & \href{FSNAU}{https://results.ipc.fsnau.org/so/} & District & 3-4 Months \\
\rowcolor[HTML]{EFEFEF} 
Remittances & \href{FSNAU}{https://fsnau.org/ids/dashboard.php} & District & Monthly \\
Credit Access & \href{FSNAU}{https://fsnau.org/ids/dashboard.php} & District & Monthly \\
\rowcolor[HTML]{EFEFEF} 
IPC & \href{FSNAU}{https://results.ipc.fsnau.org/so/} & District & 3-4 Months \\
\hline
\end{tabular}
\label{tab:data}
\end{table}

Data are harmonized at administrative level 2 (districts) across four temporal resolutions: yearly, seasonal, monthly, and IPC analysis intervals, each addressing different sources of uncertainty. IPC outcomes are normalized by district population, and covariates are standardized to a zero mean and unit variance. The treatment variable is binarized at different thresholds, with ambiguous samples near the cutoff excluded.

We estimate the Average Treatment Effect (ATE) of credit access on IPC 3+ using the back-door criterion, with the following parent adjustment set:
\[
\begin{aligned}
Z = \big(&\text{market prices, sorghum production, fatalities,} \\
          &\text{drought IDPs, population, remittances}\big).
\end{aligned}
\]
Propensity scores, trimmed to [0.30,0.80], are used as a diagnostic tool to assess covariate overlap and the plausibility of unconfoundedness, while main estimates are produced using meta-learning algorithms.

\vspace{0.5cm}

\begin{nolinenumbers}
\subsection*{\em Causal inference and effect estimation}
\end{nolinenumbers}

Although this analysis focuses on Somalia, the methodology is designed to be generalizable and replicable for food insecurity assessments across diverse global contexts.
First, it is necessary to formalize causal relationships to analyze the effects of specific variables on others. Causal inference provides a language for formalizing structural knowledge about the data-generating process via Structural Causal Models (SCMs) \citep{pearl_2009,Runge23causalreview}. With SCMs, we can estimate what will happen to data after interventions are made to its generating process. Beyond estimating interventions, SCMs also enable us to model counterfactual scenarios and assess what would have happened under different conditions. 

\begin{figure}[t]
\centering
\includegraphics[width=12 cm]{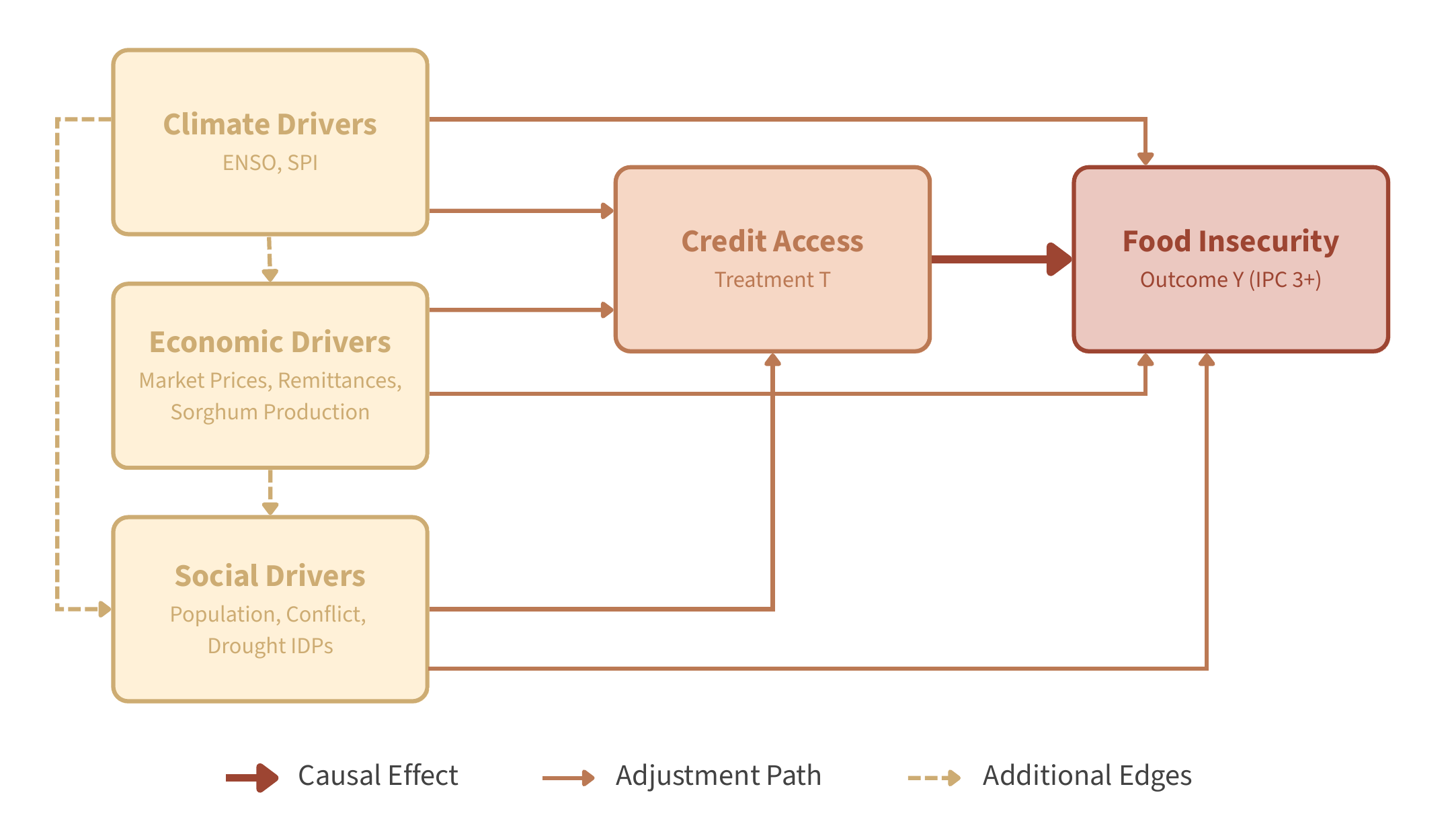}
\caption{\textbf{Directed acyclic graph (DAG) representing key determinants of food security in Somalia and their interactions.} Access to credit (a proxy for financial support) serves as the treatment variable, while IPC 3+ is the outcome of interest. The Credit Access $\rightarrow$ IPC 3+ arrow indicates the causal relationship under investigation. The directed edges illustrate hypothesized dependencies informed by domain knowledge and the literature.}
\label{fig:causal_graph}
\end{figure}

The canonical representation of causal relations is a causal Directed Acyclic Graph (DAG), which encodes a priori assumptions about the causal structure of interest. The DAG for this study is shown in Fig.~\ref{fig:causal_graph}, categorizing variables into climate, 
economic, social, and causal-effect domains; further details on its construction can be found in \textit{Appendix C} of the Supplementary Material.

\vspace{0.5cm}

\begin{nolinenumbers}
\subsection*{\em Average Treatment Effect (ATE)}
\end{nolinenumbers}

Once the DAG is established, we can explore how interventions impact the system of interest. These interventions involve a treatment variable, $T$, where $T=1$ indicates that treatment has occurred and $T=0$ indicates that it has not. Our goal is to understand the effect of $T$ on the outcome variable, $Y$. 
Nevertheless, a key challenge in causal inference is that we cannot simultaneously observe the same unit with and without treatment. To address this, we use the potential outcomes framework. For any unit $i$, $Y_{1i}$ represents the outcome if treated, and $Y_{0i}$ the outcome if untreated. The difference, $Y_{1i} - Y_{0i}$, defines the Individual Treatment Effect (ITE), but since only one of these outcomes is observable (the factual), the other remains counterfactual.  
While individual treatment effects are unobservable, we can estimate the Average Treatment Effect (ATE) across a population:
\begin{equation}
    \text{ATE} = \mathop{\mathbb{E}}[Y_{1} - Y_{0}],
\end{equation}
where $\mathop{\mathbb{E}}[\cdot]$ denotes the expected value. Estimating the ATE allows us to infer the population-level causal effect, aiding decision-making by addressing whether treatment should be applied.  

To obtain unbiased ATE estimates, we must avoid confounding bias, which arises when differences between treated and untreated populations exist before treatment. This can happen due to factors influencing treatment assignment beyond $T$ itself. These factors, known as confounders, can distort causal estimates unless properly controlled. Identifying and accounting for confounders requires a deep understanding of the system, often guided by the causal DAG.  

Finally, translating the conceptual DAG into a practical one involves mapping theoretical variables to observable proxies, a process that may not always be straightforward. This step may require domain expertise and careful consideration of assumptions, which are crucial in causal modeling when working with observational data. Experimental design choices, where feasible, can further strengthen causal claims.

\vspace{0.5cm}

\begin{nolinenumbers}
\subsection*{\em Making the right assumptions in causal inference}
\label{assumptions}
\end{nolinenumbers}

For causal inference to be valid with observational data, certain assumptions must be met \cite{Runge2021}. Ensuring these assumptions hold is crucial, as violations can introduce bias and undermine causal claims. If assumptions are not met, observed associations may misrepresent true causal relationships, emphasizing the need for careful study design, methodology, and analysis.

Within the causal inference framework, the term identifiability is a key concept that must be addressed before estimating any treatment effect. Identifiability is a statistical concept that refers to the ability of causal quantities to be uniquely inferred from the observed data. As ATE estimates require counterfactual outcomes that are not observable, formal assumptions about the data-generating process must be made to ensure identifiability \citep{pearl_2009, rubin1974}.

In addition to considering independent and identically distributed data, three assumptions are usually made in treatment effect estimation tasks \citep{imbens2015, rubin1974}. The stable unit treatment value assumption (SUTVA), also known as the consistency assumption, ensures that treatment assignment does not affect other units (no interference) and that there is no hidden heterogeneity in treatment effects. Second, the positivity assumption requires a nonzero probability of receiving the treatment for each possible combination of covariate values. Third, unconfoundedness means that, given the observed covariates, the treatment assignment is independent of potential outcomes. This assumption is fulfilled if all confounders, covariates that affect both the treatment and outcome, are observed.

Assessing the plausibility of these assumptions is crucial for the validity of treatment effect estimations. The consistency assumption is usually validated through domain expertise. Still, in our case, removing samples where the treatment is close to the binary treatment threshold helped ensure we did not violate this assumption. To verify the positivity assumption, we typically examine the overlap of propensity scores, which represent the probability of receiving treatment based on covariates \citep{rubin1974}, of treated and untreated populations (see \emph{Appendix B}). 
Propensity-score overlap is a diagnostic for the positivity assumption. We assess overlap empirically (\emph{Appendix B}) and apply trimming to reduce the influence of extreme estimated probabilities. We note that overlap patterns depend on the propensity model and should be interpreted as a diagnostic rather than a guarantee of identification.

Validating the assumption of unconfoundedness is especially difficult with real-world data, and the best way to ensure it is to consult experts in the field and draw on their domain expertise. It is worth noting that this assumption may be challenged in our context due to limitations in data coverage, measurement error, and potential endogeneity. In particular, some explanatory variables (e.g., food prices, sorghum production) may also be indirectly used in constructing the IPC outcome variable. Although the DAG-based design and robustness checks described below help mitigate these risks, we interpret estimated effects as approximate treatment effects within a quasi-experimental framework rather than definitive causal quantities under strict ignorability.

\vspace{0.5cm}

\begin{nolinenumbers}
\subsection*{\em ATE estimation methods}
\end{nolinenumbers}

Using the Potential Outcomes framework, we estimate the ATE by comparing IPC 3+ values when access to credit exceeds a predefined threshold with those when it remains below that threshold. To estimate the effect, we use several methods of varying complexity. Linear regression (LR) and propensity score matching (M) are selected as baseline estimation methods. The popular Inverse Propensity Score weighting (IPS W) is also used \citep{stuart}, as well as modern machine learning methods, the $T$-Learner (T-L) and $X$-Learner (X-L) \citep{kunzel}, known as meta-learner methods. In what follows, we focus on the $X$-Learner, and in {\em Appendix~D}, we provide the estimation for all methods used. It is worth noting that the $X$-Learner outperforms the other methods in this context.

Meta-learners are machine learning frameworks that estimate treatment effects using existing supervised learning algorithms. These methods are flexible and modular, allowing the choice of base models that best fit the data and domain. The meta-learners are agnostic to the choice of the base learner, and each stage (e.g., outcome prediction, propensity score estimation) can be fine-tuned separately. 

The $X$-Learner is particularly useful in settings with imbalanced treatment groups. It uses cross-fitting to better handle differences in sample size. The $X$-Learner works by initially training two models $f_1$ and $f_0$ to predict outcomes for treated $\hat{Y}_{1}(X)$ and control $\hat{Y}_{0}(X)$ groups, respectively. The second step consists of computing pseudo-treatment effects for each group: for treated individuals, $D_1 = Y - \hat{Y}_{0}(X)$, and for control individuals, $D_0 = \hat{Y}_{1}(X) - Y$. The next step is to train models $g_1$ and $g_0$ to predict $D_1$ and $D_0$ from the features $X$. Finally, the treatment effect is obtained by combining the predictions. If the treated group is smaller, then
\begin{equation}
    \hat{\tau}(X) = p(X)g_1(X) + (1-p(X))g_0(X),
\end{equation}
where $p(X)$ is the propensity score, estimating the likelihood of treatment.

\vspace{0.5cm}

\begin{nolinenumbers}
\subsection*{\em On results validation and assessing robustness}
\label{refutation_tests}
\end{nolinenumbers}

Statistical significance is essential for drawing reliable conclusions from sample data, helping to distinguish genuine effects from random variation. High-quality data is crucial for ensuring the validity of statistical tests used in causal estimation. 
Given the lack of ground truth estimates, we conduct refutation tests to evaluate the robustness of our models, following recent research \citep{sharma2020dowhy, cinelli2019}. 
These tests assess whether the estimated causal relationships stem from true causation rather than confounding, selection bias, or data inconsistencies. The results confirm that statistically significant estimates are not artifacts of data errors but instead reflect genuine causal effects, as validated across multiple refutation tests. Therefore, while individual variables may shift the magnitude of the estimated effect, the direction and statistical significance remain stable across specifications, minimizing the risk of omitted variable bias.

We perform the following tests: i) Placebo test, where we randomly permute the treatment variable, expecting the estimated effect to converge to zero; ii) Random Common Cause (RCC), where a random confounder is added to the dataset and the estimate is expected to remain unchanged; iii) Random Subset Removal (RSR), where a subset of data is randomly removed, and the effect is expected to remain the same. Experimental results with the different refutation tests can be found in \textit{Appendix E} of the Supplementary Material.

\vspace{0.5cm}

%=======================================================
% 3. RESULTS
%=======================================================

\begin{nolinenumbers}
\section*{Results}
\label{sec:results}
\end{nolinenumbers}

\begin{nolinenumbers}
\subsection*{\em Credit access is associated with lower acute food insecurity in Somalia}
\end{nolinenumbers}

\begin{figure}[t]
\centering
\leftskip-1cm
\includegraphics[width=1.2\textwidth]{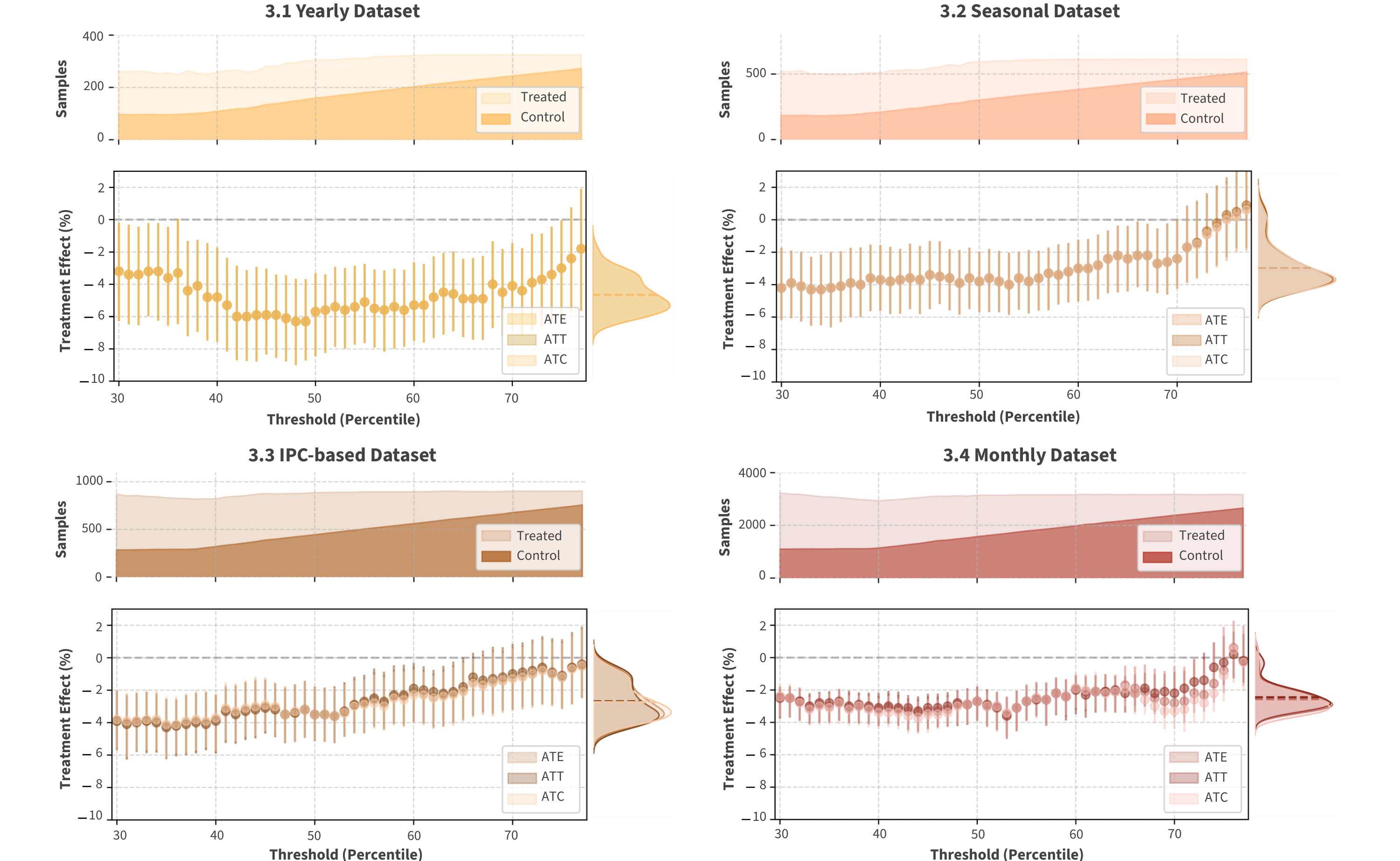}
\vspace{0.1cm}
\caption{\textbf{Treatment Effect of Credit Access on IPC 3+ populations across different data aggregations: yearly, seasonal, IPC-based, and monthly.}
Each panel represents a distinct aggregation level, showing ATE, ATT, and ATC estimates across treatment thresholds. The top section of each panel displays the distribution of treated- and control-group samples, while the bottom section shows treatment estimates with confidence intervals. The density plots on the right depict the distribution of ATE, ATT, and ATC values. These results highlight how treatment effects vary depending on data aggregation and treatment definition.}
\label{fig:ate_results_meta}
\end{figure}

Using the Potential Outcomes Framework \cite{rubin1974}, we estimate the impact of higher credit access on the percentage of the population in IPC Phase 3+, under the identification assumptions described in the \emph{Methods} section. We estimate the impact via the Average Treatment Effect (ATE) at the district level in Somalia across four temporal aggregations: yearly, seasonal, monthly, and intervals between IPC analyses (IPC-based). We choose these aggregations to address suboptimal data quality and quantity. With the yearly or seasonal aggregations, we aim to avoid unknown data inconsistencies by averaging the values per year or rainy season. In contrast, with the monthly or IPC-based aggregations, we prioritize data quantity by maximizing the total or outcome sample counts in the experiments, respectively (see \textit{Methods} subsection \textit{Data collection and setup}). A negative ATE indicates that the treated group, those with sufficient credit access, has fewer people in IPC Phases 3 or worse than the control group. Statistically significant results show an average improvement between 2\% and 4\% in the treated subpopulation across the different temporal aggregations (see Fig.~\ref{fig:ate_results_meta}). 

We present results for different estimation methods and treatment binarizations (percentile thresholds), which can be understood as different levels of the strictness of treatment definition: low percentiles refer to a relaxed definition, where a higher number of samples have received enough credit to be considered as treated, and vice versa for high percentiles, where we imply a stricter definition of the treatment. We employ a binary treatment to account for the non-linear relationship between credit access and IPC changes and to mitigate noise in the continuous proxy. The ATE estimate, therefore, quantifies the average difference in the proportion of individuals classified in IPC Phases 3+ between populations that receive enough credits and those that do not. This estimate also aligns with operational policy relevance by assessing the impact when transfers exceed a population-adjusted threshold. More details on the methods and data preprocessing can be found in the \textit{Methods} section, specifically in the \textit{Average Treatment Effect (ATE)} and \textit{Data collection and setup} subsections. Robustness is confirmed across all specifications via the refutation tests described in the \textit{Methods} subsection \textit{On results validation and assessing robustness}; full results appear in \textit{Appendix E} of the Supplementary Material.

We also conducted a comparative analysis of different estimation metrics. In causal-effect estimation, it is often useful to assess the treatment effect on either the treated or control group, rather than relying solely on the overall average effect. These subgroup estimates are known as the Average Treatment Effect on the Treated (ATT) and the Average Treatment Effect on the Control (ATC). Our comparison of ATE, ATT, and ATC revealed no significant differences, indicating that the treatment effect is evenly distributed across subgroups. Such similarity suggests minimal heterogeneity in treatment effects between groups.

\vspace{0.5cm}

\begin{nolinenumbers}
\subsection*{\em 
Benefits Persist Despite Confounding Influences of Prices, Yields, and Displacement}
\end{nolinenumbers}

\begin{figure}[t]
    \centering
    {\includegraphics[width=1\textwidth]{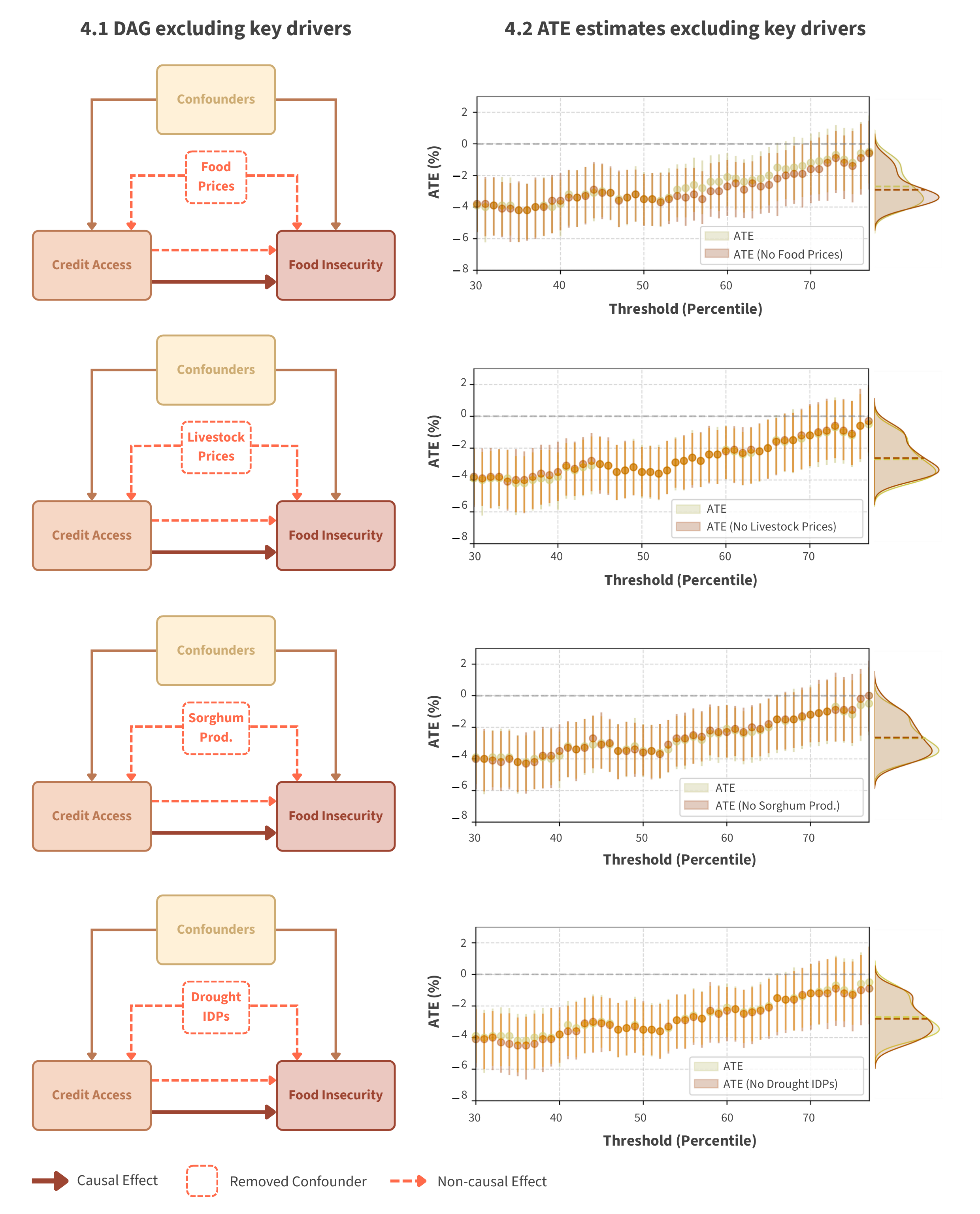}}
    \caption{
    {\bf Sensitivity analysis of confounders in causal effect estimation.} 
    The left column illustrates causal diagrams excluding a key confounder, 
    while the right column shows the corresponding ATE estimates.
    }
    \label{fig:sensitivity_results_meta}
\end{figure}

To evaluate the robustness of our causal estimates, we conducted a sensitivity analysis focusing on other key factors, also known as confounding variables \cite{robins2000sensitivity}. Using a {\em leave-one-out (LOO)} approach, we systematically removed individual confounders (food prices, livestock prices, sorghum yields, and the number of drought-related internally displaced persons) to assess how each one influences the estimated effect of access to credit on food security outcomes. 

The analysis, using the IPC-based aggregation data, reveals small shifts in estimated treatment effects when individual covariates are removed from the model (see Fig.~\ref{fig:sensitivity_results_meta}). These variations are not statistically significant (see \textit{Appendix D} of the Supplementary Material), and confidence intervals overlap substantially; no single variable dominates the estimation, and the overall causal signal is stable across covariate configurations. Therefore, the leave-one-out results support the robustness of the estimated credit-access effect to plausible perturbations of the adjustment set.

\vspace{0.5cm}

\begin{nolinenumbers}
\subsection*{\em Significant Gains Despite the Unstructured Heterogeneous Transfer Distribution} 
\end{nolinenumbers}

To assess overlap and support the plausibility of the positivity assumption, we compute propensity scores prior to effect estimation (\emph{Appendix B}), following the diagnostic approach described in the \textit{Methods} section.
Given the convergence of the ATE, ATT, and ATC values (see Fig.~\ref{fig:ate_results_meta}), we observe that the treated and control groups have similar average covariate values. When both groups are balanced in baseline characteristics, treatment effect differences between subgroups diminish, as neither group is disproportionately likely to receive treatment. The propensity scores outcome indicates sufficient overlap between treated and untreated units, suggesting that units with similar observed covariates received different treatments. This overlap indicates a lack of strong systematic bias in the covariates and supports causal inference under observational conditions, thereby minimizing systematic differences in observed outcomes between treated and untreated groups.
Given limited metadata on the credit measure (e.g., targeting criteria, implementing actors, and transfer design), we cannot characterize the allocation mechanism as targeted or untargeted. We therefore avoid strong claims about program design and focus on estimating effects conditional on observed covariates, supplemented by robustness and refutation tests.

\vspace{0.5cm}

%=======================================================
% 4. DISCUSSION
%=======================================================

\begin{nolinenumbers}
\section*{Discussion}
\end{nolinenumbers}

This study provides observational evidence that higher reported access to credit is associated with lower acute food insecurity in Somalia, under explicit identification assumptions, offering insight into how financial interventions can be scaled across diverse global food system contexts. Our findings are particularly relevant for countries facing similar constraints in targeting, data availability, or operational infrastructure.

Using observational data from a range of environmental, economic, and social sources, we find that credit access is linked with consistent improvements in food security outcomes, even in the absence of organized access to credit. 
Specifically, across all districts and years analyzed, we estimate that access to credit reduces the risk of acute food insecurity by 2\% when measured over the total population. While this effect may seem modest in absolute terms, it is more substantial when considered relative to the population at risk. On average, 16\% of the population in the studied districts is experiencing crisis-level food insecurity, meaning that a 2\% absolute reduction represents a more meaningful shift within this vulnerable group. 

Importantly, these conclusions are derived using causal inference methods applied to observational data, thereby constituting a sound, quantitative, and agnostic approach to the problem. Yet, at the same time, causal inference introduces certain methodological constraints \cite{Runge2021, Runge23causalreview}. Because experimental or randomized designs are not feasible, our estimates rely on assumptions regarding the relationships among variables and how we observe them. While we carefully constructed a causal model and included a broad set of covariates, we acknowledge the possibility that unobserved or poorly measured factors may still bias the results.  
Additionally, gaps in the credit access data, such as incomplete documentation of timing, targeting criteria, and implementation quality, limit our ability to precisely attribute effects.

Our primary outcome variable, IPC, also has inherent limitations. While IPC is the de facto standard for guiding food security responses in Somalia and similar contexts, the classification process involves expert judgment that may vary with data availability, committee composition, and contextual factors. This introduces potential opacity and inconsistency across space and time, which limits IPC's interpretability as a strictly observational outcome and can induce non-random measurement variability. This variability may attenuate the estimated treatment effects and should be considered when interpreting the magnitude and robustness of our findings.

We acknowledge this limitation and note that, while IPC is well-suited for policy relevance, future work could benefit from access to the underlying household-level indicators used in its construction or from more precise measures such as the People in Need (PiN) index \citep{PINHA2023}. However, IPC remains the most widely used food insecurity metric in Somalia and the standard for operational decision-making. Its availability at the district level and integration of multiple indicators into a structured classification make it especially suitable for evaluating crisis responses in data-constrained contexts like this one.
It is also worth noting that, despite the limitations mentioned, the consistency of treatment effects across multiple robustness checks, combined with statistically significant estimates and adequate propensity score overlap, lends confidence in the internal validity of our findings.

Another key concern in this study is the assumption that all factors affecting both treatment assignment and outcomes are accounted for. To assess the robustness of this assumption, we conducted refutation tests to evaluate potential biases and validate the model's assumptions. We also conducted sensitivity analyses, including LOO tests, to examine the influence of individual covariates. While these tests help gauge the reliability of our estimates, the potential for hidden bias remains, particularly given the complexity and fluidity of humanitarian contexts.

Additionally, we classified districts as either treated or untreated based on percentile thresholds of credit access coverage, effectively binarizing what is, in reality, a continuous and uneven intervention.
This may overlook important nuances, such as differences in transfer amounts and timing, and in whether (and how) individual households received the transfers. While this approach facilitates more precise estimation, it may also mask variations in the intervention's true impact.

\begin{figure}[t!]
\begin{center}
\leftskip0.6cm
\includegraphics[width=12cm]{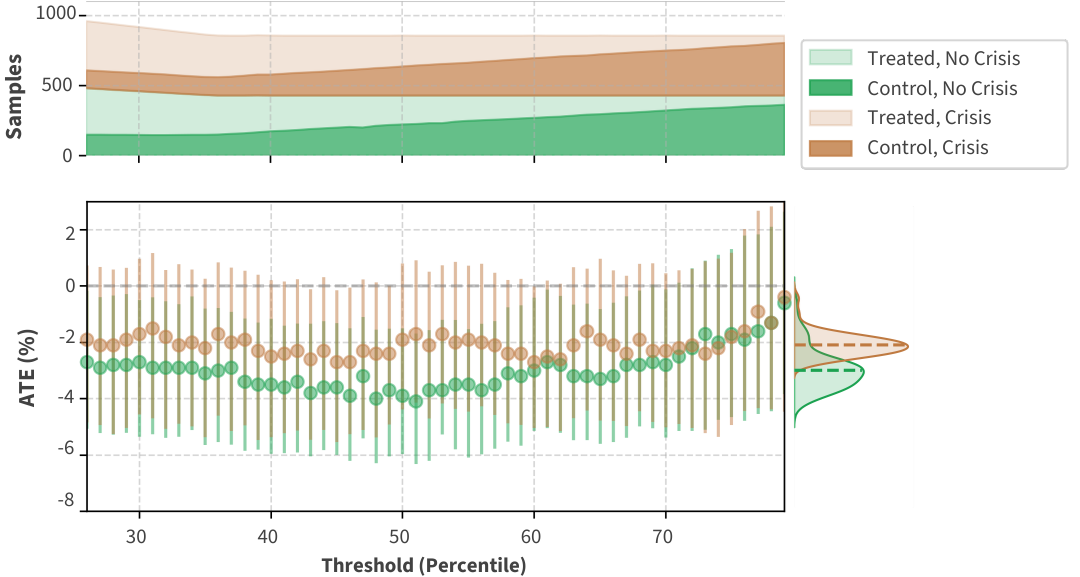}  
\vspace{0.1cm}
\end{center}
\caption{\textbf{Heterogeneity analysis of the causal effect of credit access on IPC 3+ populations under two crisis scenarios using the IPC-based data aggregation.} 
The top panel shows the distribution of treated and control group samples across different treatment thresholds, distinguishing between crisis and non-crisis conditions. The bottom panel presents the estimated Average Treatment Effect (ATE) for each threshold and crisis scenario, along with the distribution of all ATE estimations.}
\label{fig:heterogeneity_analysis}
\end{figure}

Despite these technical limitations, the policy implications are clear. Even when cash assistance is not rigorously targeted, it can still deliver noticeable improvements in food security. Moreover, it also suggests significant room for enhancement, as more deliberate and data-informed targeting could increase the observed benefits \citep{Arnold2025}. For example, with enough data, we could investigate the impact of credit access on food security indices under different vulnerability conditions (i.e., higher numbers of conflicts and internal migrations; see Fig.~\ref{fig:heterogeneity_analysis}). While these results show variations in ATE estimates, they are not statistically significant with the available data. A more granular approach could provide deeper insights. Instead of estimating a single ATE for the entire country, we could leverage Conditional Average Treatment Effect (CATE) estimates to analyze how access to credit varies across different regions. 
However, this finer-grained analysis requires a larger sample to ensure statistical significance.

The findings also highlight the need for better-integrated data systems that combine near real-time environmental indicators with up-to-date socioeconomic and programmatic data \citep{Porciello2020}. Such systems could support more agile humanitarian responses and provide decision-makers with stronger evidence to design or adapt aid programs. This, in turn, could contribute to improved resilience and long-term development outcomes for communities affected by recurring food crises. While our analysis covers the majority of Somalia's population, it is based on 56 of the country's 72 districts, due to data completeness constraints. While no structural bias was detected, we acknowledge that data limitations, particularly in more insecure or inaccessible districts, may affect the spatial generalizability of the results. These gaps underscore a key barrier to scaling this kind of analysis in fragile settings: the need for consistent, high-resolution observational data. Nonetheless, the method's flexibility enables its use even with partial data availability, making it relevant to other regions facing similar structural constraints.

Beyond the regional focus, our framework can be adapted to assess food policy interventions in other fragile settings. As global institutions seek to scale evidence-based strategies across different food-insecure contexts, this approach offers a transferable model for evaluating impact and improving response design.

\vspace{0.5cm}

%=======================================================
% 5. CONCLUSION
%=======================================================

\begin{nolinenumbers}
\section*{Conclusion}
\end{nolinenumbers}

This study demonstrates the potential of access to credit to mitigate acute food insecurity, even in challenging and data-scarce settings. Applying an observational causal machine-learning framework to a harmonized dataset spanning environmental, socioeconomic, and conflict-related variables from 2015 to 2022, we find that higher credit access is associated with a 2\% reduction in the proportion of the population in IPC Phase 3+ across districts in Somalia. Given that an average of 16\% of the population is in crisis-level food insecurity, this absolute reduction represents a meaningful shift within the at-risk group.

The consistency of this finding across temporal aggregations, estimation methods, and robustness checks, including placebo tests, random common cause tests, and leave-one-out sensitivity analyses, lends credibility to the estimated effect, despite the inherent limitations of observational data. Our results suggest that even in the absence of rigorously targeted allocation, financial access channels can yield tangible improvements in food security outcomes.

Moving forward, enhancing data quality, particularly in treatment allocation and socioeconomic indicators, will be critical to strengthening causal estimates \citep{NatureFood2023}. Expanding the use of advanced causal methodologies can further refine impact assessments, supporting evidence-based decision-making in resource-constrained and crisis-affected scenarios. By bridging data-driven insights with policy action, future efforts can optimize humanitarian interventions and enhance resilience against food crises in vulnerable regions.

\clearpage

\begin{nolinenumbers}
\section*{Data availability} 
\end{nolinenumbers}

All datasets used in this study are available through open-access sources (see Table~\ref{tab:data}). The Harmonized Food Security Data, compiled specifically for this study, will be accessible via this work's GitHub repository \href{Causal4FS-Credit}{https://github.com/jordicbau/Causal4FS-Credit}. The data preprocessing code used for harmonization and analysis will also be provided in the same repository. 

\begin{nolinenumbers}
\section*{Code availability} 
\end{nolinenumbers}

All code used for the analysis in this paper will be openly available on GitHub at \href{Causal4FS-Credit}{https://github.com/jordicbau/Causal4FS-Credit}. The effect estimation analysis was conducted using the open-source Python packages \href{CausalML}{https://github.com/uber/causalml} and \href{DoWhy}{https://github.com/py-why/dowhy}.
\newpage

\begin{nolinenumbers}
\bibliographystyle{unsrtnat}
\bibliography{ref}

\clearpage

\section*{Acknowledgments} 

This work was supported by the Fundaci\'on BBVA with the project Causal inference in the human biosphere coupled system (\href{SCALE}{https://www.fbbva.es/noticias/concedidas-5-ayudas-a-equipos-de-investigacion-cientifica-en-big-data/}), the Microsoft Climate Research Initiative through the \href{Causal4Africa}{https://www.microsoft.com/en-us/research/collaboration/microsoft-climate-research-initiative/projects/} project, and the European Union's Horizon Europe Research and Innovation Program through the \href{ThinkingEarth}{https://thinking-earth.eu/} project (under Grant Agreement number 101130544). GCV research for this study was funded by the European Research Council (ERC) Synergy Grant ``Understanding and Modeling the Earth System with Machine Learning'' (USMILE) under the Horizon 2020 Research and Innovation program (Grant Agreement No. 855187).

\section*{Author contributions} 

J.C-B., V.S., and G.C-V. designed and structured the study. J.C-B. and G.C-V. wrote the initial draft, with contributions and revisions from all co-authors.
J.M.L-P. provided expertise on the definition of IPC and the implications of key assumptions. D.P. contributed to discussions on the study's impact on humanitarian response.
J.C-B. led data harmonization, curation, and collection, with early support from J.M.T., who also integrated migration and internal displacement data. %
G.C-V. supervised the study from inception and secured funding.
All authors contributed to the manuscript and reviewed its final version.

\section*{Competing interests} 

The authors declare no competing interests.

\end{nolinenumbers}

\newpage

\begin{nolinenumbers}
\section*{Supplementary information}
\end{nolinenumbers}

\begin{appendices}

% ============================================================
\begin{nolinenumbers}
\section{Study districts comparison}\label{app:districts}
\end{nolinenumbers}

\begin{figure}[H]
\centering
\includegraphics[width=0.8\textwidth]{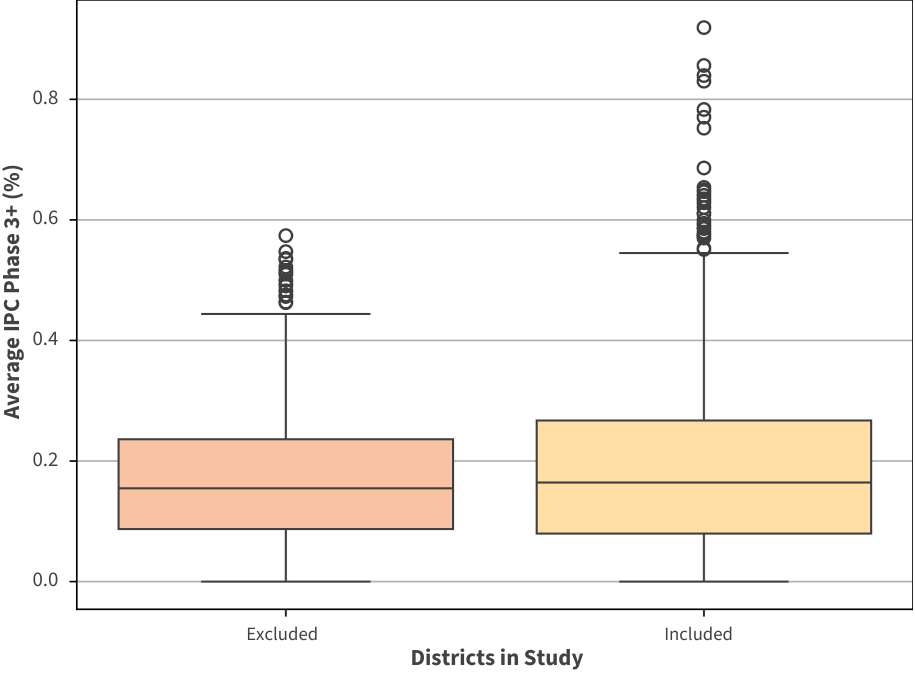}
\caption{\textbf{Comparison of food insecurity levels in included vs.\ excluded districts.}
IPC 3+ values are normalized by district population. Blue: included districts (n=56). Orange: excluded districts (n=16).}
\label{fig:districts_study}
\end{figure}

Figure~\ref{fig:districts_study} compares the proportion of the population classified as IPC Phase 3+ (crisis-level food insecurity or worse) between districts included in the analysis and those excluded due to incomplete data. The IPC 3+ percentage was normalized by district population to ensure comparability. Districts labeled as \textit{included} are those with complete and consistent data across all variables used in the causal analysis, whereas \textit{excluded} districts lacked key indicators (e.g., IPC phase classifications, covariates, or treatment data) over the study period (2015--2022). The number of districts in each group is shown above the corresponding boxplot. Overall, the distributions are similar, suggesting that the included districts remain broadly representative of observed IPC 3+ levels. Nevertheless, we acknowledge that data availability may induce selection effects; these limitations are discussed in the main manuscript.

% ============================================================
\begin{nolinenumbers}
\section{Propensity score calculation}\label{app:propscores}
\end{nolinenumbers}

\begin{figure}[H]
\centering
\includegraphics[width=0.9\textwidth]{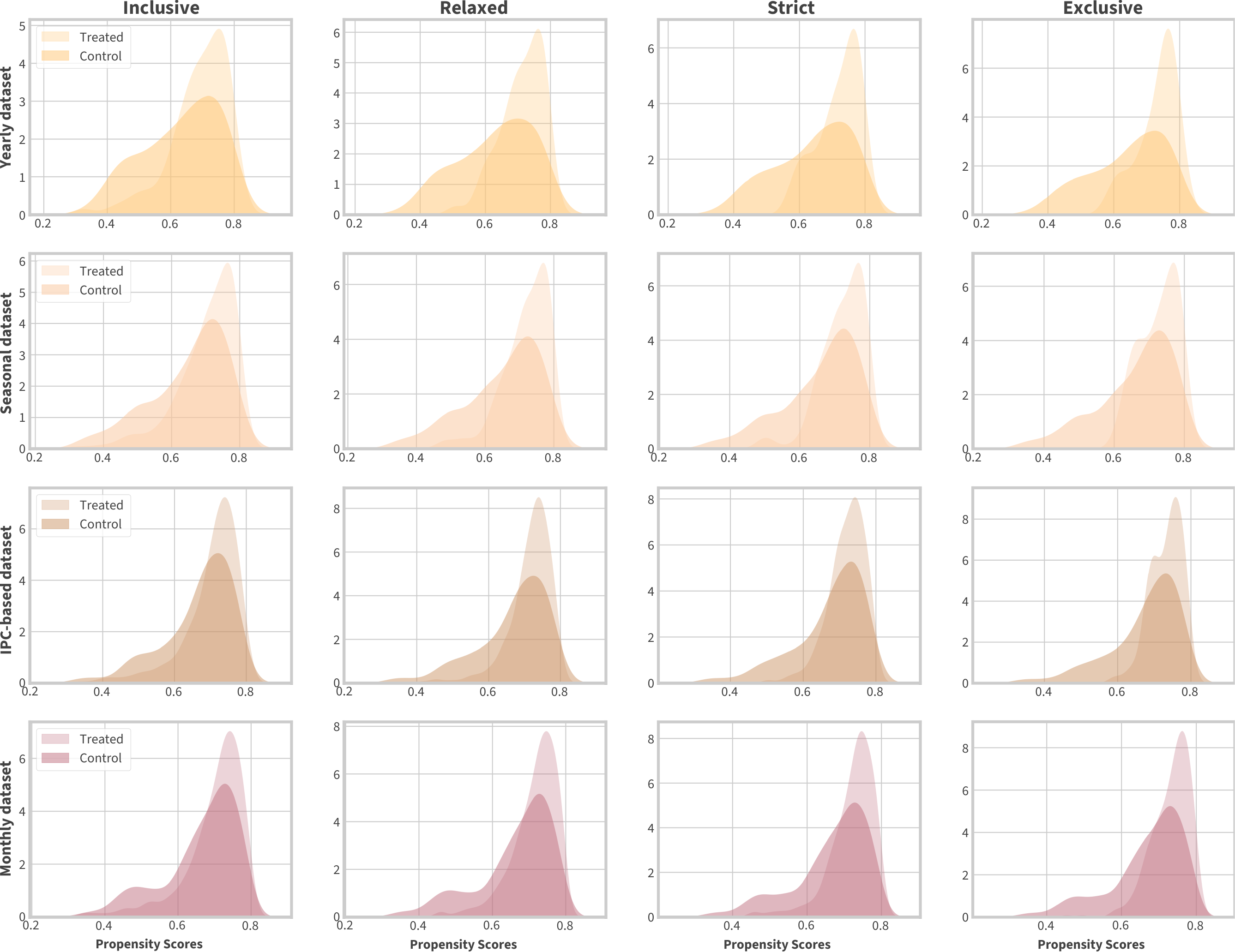}
\caption{\textbf{Kernel density estimates of propensity scores across four data aggregations.}
Rows correspond to data aggregation (yearly, seasonal, monthly, IPC-based). Columns correspond to treatment thresholds (30th/45th/60th/75th percentiles), with increasing strictness from left to right.}
\label{fig:propensity_scores}
\end{figure}

Figure~\ref{fig:propensity_scores} shows estimated propensity score distributions across the four data aggregations (yearly, seasonal, monthly, and IPC-based), under four increasing levels of treatment strictness: \textit{inclusive} (30th percentile), \textit{relaxed} (45th percentile), \textit{strict} (60th percentile), and \textit{exclusive} (75th percentile). The overlap between treated and control densities provides empirical support for the feasibility of effect estimation under the overlap assumption. At the same time, the degree of separation between densities reflects how strongly the observed covariates predict treatment assignment. In a randomized experiment, treatment is independent of covariates and the true propensity score is constant (equal to the marginal treatment probability); in observational settings, estimated propensity scores often vary substantially and need not be centered at 0.5.

\newpage

% ============================================================
\begin{nolinenumbers}
\section{Justification of causal links in the DAG}\label{app:dag_links}
\end{nolinenumbers}

To support the structure of the directed acyclic graph (DAG) used in our causal analysis, we provide a literature-based justification for each causal link. The DAG encodes hypothesized relationships among environmental, socioeconomic, and demographic variables that plausibly influence both the receipt of credit access and IPC 3+ outcomes. Table~\ref{tab:dag_justification} summarizes the directional assumptions and references supporting each link (peer-reviewed literature and technical sources, including the IPC Global Technical Manual, FAO, and related food security reports). These justifications ground the adjustment strategy in domain knowledge and guide the identification of plausible confounder sets.

\begin{table}[h]
\centering
\leftskip-1.0cm
\small
\caption{{\bf Justification of causal links in the DAG based on literature.}}
\begin{tabular}{p{3.6cm}p{6.3cm}p{2.8cm}}
\hline
\textbf{Causal Link} & \textbf{Justification} & \textbf{Reference(s)} \\
\hline
\rowcolor[HTML]{EFEFEF}
ENSO $\rightarrow$ SPI & ENSO modulates rainfall anomalies, impacting drought frequency. &
\makecell[l]{Gitz et al. (2016)\\FAO (2023)} \\
ENSO $\rightarrow$ Market prices; Sorghum Production & Climatic shocks influence agricultural yields and market prices. &
\makecell[l]{Gitz et al. (2016)\\FAO (2023)} \\
\rowcolor[HTML]{EFEFEF}
SPI $\rightarrow$ Market prices; Sorghum Production & Rainfall deficits reduce production and raise prices. &
\makecell[l]{Misselhorn (2005)\\iDMC (2020)} \\
SPI $\rightarrow$ Remittances & Drought increases remittance reliance as coping. &
\makecell[l]{Misselhorn (2005)\\FSIN (2020)} \\
\rowcolor[HTML]{EFEFEF}
Market prices; Sorghum Production $\rightarrow$ Fatalities, Drought IDPs & Resource stress may increase conflict and displacement. &
\makecell[l]{Clapp et al. (2022)\\FSIN (2020)\\iDMC (2020)} \\
Fatalities, Drought IDPs, Population, Remittances $\rightarrow$ Credit Access & Humanitarian targeting and household responses are triggered by distress signals. &
\makecell[l]{IPC Manual (2021)\\FAO (2023)\\iDMC (2020)} \\
\rowcolor[HTML]{EFEFEF}
Population $\rightarrow$ Drought IDPs, Remittances, treatment & Larger populations face greater exposure and migration potential. &
\makecell[l]{CFS (2012)\\FAO (2023)\\iDMC (2020)} \\
Remittances $\rightarrow$ Market prices, Credit Access, IPC 3+ & Remittances affect demand, aid likelihood, and food security. &
\makecell[l]{Fyles et al. (2016)\\Misselhorn (2005)} \\
\rowcolor[HTML]{EFEFEF}
All stressors $\rightarrow$ IPC 3+ & Combined shocks drive acute food insecurity. &
\makecell[l]{IPC Manual (2021)\\FSIN (2020)\\iDMC (2020)} \\
\hline
\end{tabular}
\label{tab:dag_justification}
\end{table}

\newpage

% ============================================================
\begin{nolinenumbers}
\section{\textit{Leave-One-Out} significance tests}\label{app:loo_tests}
\end{nolinenumbers}

To assess the robustness of our estimated ATE to the assumed adjustment set, we conducted a leave-one-out (LOO) sensitivity analysis. We re-estimated the ATE after sequentially removing each key confounder (food prices, livestock prices, sorghum production, and drought-induced internal displacement) from the adjustment set. Table~\ref{tab:loo_summary} reports the mean difference in ATE between the full model and each LOO model, the 95\% bootstrap confidence interval of the ATE difference, and the p-value for testing whether this difference differs from zero. All estimates are based on bootstrap distributions ($n=1000$) per threshold (percentiles 30--75), and are aggregated over thresholds. Across all dropped covariates, the differences are not statistically significant, suggesting that the estimated effect is not driven by a single confounder alone.

\begin{table}[h]
\centering
\small
\caption{{\bf Average difference in ATE between the full and LOO models across thresholds.}}
\begin{tabular}{lccc}
\hline
Confounder removed & Mean $\Delta$ATE & 95\% CI of $\Delta$ATE & p-value \\
\hline
\rowcolor[HTML]{EFEFEF}
Food prices & -0.002 & [-0.003, -0.001] & 0.549 \\
Livestock prices & 0.001 & [0.000, 0.002] & 0.561 \\
\rowcolor[HTML]{EFEFEF}
Sorghum Production & 0.001 & [0.000, 0.002] & 0.527 \\
Drought IDPs & -0.001 & [-0.002, -0.001] & 0.525 \\
\hline
\end{tabular}
\label{tab:loo_summary}
\end{table}

\newpage

% ============================================================
\begin{nolinenumbers}
\section{Additional experimental results}\label{app:moreresults}
\end{nolinenumbers}

Tables~\ref{tab:yr_ate_results}--\ref{tab:m_ate_results} report intermediate results across temporal aggregations. We compare ATE estimates from methods of increasing complexity: Linear Regression (LR), propensity score Matching (M), Inverse Propensity Score Weighting (IPS W), and the meta-learners $T$-Learner (T-L) and $X$-Learner (X-L). For each threshold, we report ATE estimates with 95\% confidence intervals (CI) and p-values, alongside placebo, random common cause (RCC), and random subset removal (RSR) refutation tests. Following the treatment binarization strategy used throughout the paper, we consider four treatment thresholds with increasing strictness: inclusive (30th percentile), relaxed (45th percentile), strict (60th percentile), and exclusive (75th percentile). Overall, the meta-learners tend to yield more stable estimates under treated/control imbalance; given its design advantages in imbalanced settings, we use the $X$-Learner as the primary estimator in the main experiments.

% -------------------------------
% Yearly dataset (NEW)
\clearrow
\begin{table*}[h]
    \centering
    \leftskip-1.5cm
    \caption{{\bf Treatment threshold, method, ATE estimates, 95\% confidence intervals and p-values. Refutation tests fail if their p-value is less than 0.05. Yearly dataset.}}
    \footnotesize
    \begin{tabular}{>{\rowmac}c>{\rowmac}c>{\rowmac}c>{\rowmac}c>{\rowmac}c>{\rowmac}c>{\rowmac}c>{\rowmac}c>{\rowmac}c>{\rowmac}c>{\rowmac}c<{\clearrow}}
    \hline
        \multicolumn{5}{c}{\multirow{2}{*}{Cause Effect Estimation}} & \multicolumn{6}{c}{Refutation Tests} \\ \cline{6-11}
        \multicolumn{5}{c}{} & \multicolumn{2}{c}{Placebo} & \multicolumn{2}{c}{RCC} & \multicolumn{2}{c}{RSR} \\ \hline
        Th & Method & ATE & CI & p-value & Effect* & p-value & Effect* & p-value & Effect* & p-value \\\hline
        \setrow{\bfseries} 30 & LR & -0.038 & (-0.068, -0.010) & 0.014 & 0.000 & 0.008 & -0.038 & 0.988 & -0.038 & 0.947 \\
        \rowcolor[HTML]{EFEFEF} 30 & M & -0.009 & (-0.041, 0.010) & 0.518 & 0.006 & 0.563 & -0.019 & 0.052 & -0.013 & 0.630 \\
        30 & IPS W & -0.024 & (-0.087, 0.036) & 0.447 & 0.001 & 0.125 & -0.029 & 0.030 & -0.029 & 0.481 \\
        \rowcolor[HTML]{EFEFEF} \setrow{\bfseries} 30 & T-L & -0.038 & (-0.058, -0.016) & 0.019 & 0.001 & 0.220 & -0.039 & 0.318 & -0.039 & 0.866 \\
        \setrow{\bfseries} 30 & X-L & -0.039 & (-0.056, -0.020) & 0.001 & 0.001 & 0.220 & -0.039 & 0.318 & -0.039 & 0.866 \\
        \rowcolor[HTML]{EFEFEF} \setrow{\bfseries} 45 & LR & -0.042 & (-0.074, -0.014) & 0.005 & -0.000 & 0.001 & -0.042 & 0.958 & -0.042 & 0.982 \\
        \setrow{\bfseries} 45 & M & -0.030 & (-0.060, -0.006) & 0.031 & 0.005 & 0.056 & -0.035 & 0.339 & -0.032 & 0.810 \\
        \rowcolor[HTML]{EFEFEF} 45 & IPS W & -0.011 & (-0.074, 0.052) & 0.725 & 0.001 & 0.441 & -0.038 & 0.000 & -0.037 & 0.004 \\
        \setrow{\bfseries} 45 & T-L & -0.029 & (-0.046, -0.011) & 0.004 & 0.000 & 0.200 & -0.029 & 0.771 & -0.029 & 0.963 \\
        \rowcolor[HTML]{EFEFEF} \setrow{\bfseries} 45 & X-L & -0.030 & (-0.046, -0.013) & 0.001 & 0.000 & 0.200 & -0.030 & 0.771 & -0.030 & 0.963 \\
        60 & LR & -0.019 & (-0.053, 0.012) & 0.240 & -0.000 & 0.228 & -0.019 & 0.972 & -0.019 & 0.982 \\
        \rowcolor[HTML]{EFEFEF} 60 & M & -0.028 & (-0.059, 0.007) & 0.095 & 0.001 & 0.109 & -0.028 & 0.968 & -0.027 & 0.940 \\
        60 & IPS W & -0.022 & (-0.073, 0.028) & 0.401 & -0.000 & 0.142 & -0.019 & 0.069 & -0.019 & 0.673 \\
        \rowcolor[HTML]{EFEFEF} \setrow{\bfseries} 60 & T-L & -0.020 & (-0.039, -0.003) & 0.062 & -0.000 & 0.560 & -0.033 & 0.420 & -0.033 & 0.872 \\
        \setrow{\bfseries} 60 & X-L & -0.021 & (-0.040, -0.006) & 0.016 & -0.000 & 0.560 & -0.021 & 0.420 & -0.021 & 0.872 \\
        \rowcolor[HTML]{EFEFEF} 75 & LR & -0.014 & (-0.101, 0.027) & 0.684 & -0.001 & 0.563 & -0.014 & 0.968 & -0.022 & 0.927 \\
        75 & M & 0.001 & (-0.055, 0.032) & 0.967 & -0.004 & 0.974 & -0.009 & 0.182 & -0.005 & 0.632 \\
        \rowcolor[HTML]{EFEFEF} 75 & IPS W & 0.001 & (-0.075, 0.084) & 0.989 & -0.002 & 0.991 & 0.005 & 0.135 & 0.004 & 0.788 \\
        75 & T-L & -0.014 & (-0.035, 0.012) & 0.640 & -0.000 & 0.608 & -0.009 & 0.098 & -0.009 & 0.729 \\
        \rowcolor[HTML]{EFEFEF} 75 & X-L & -0.012 & (-0.032, 0.010) & 0.259 & -0.000 & 0.608 & -0.004 & 0.098 & -0.006 & 0.729 \\
        \hline
    \end{tabular}
    \label{tab:yr_ate_results}
\end{table*}

\clearrow
\begin{table*}[h]
    \centering
    \leftskip-1.5cm
    \caption{{\bf Treatment threshold, method, ATE estimates, 95\% confidence intervals and p-values. Refutation tests fail if their p-value is less than 0.05. Seasonal dataset.}}
    \footnotesize
    \begin{tabular}{>{\rowmac}c>{\rowmac}c>{\rowmac}c>{\rowmac}c>{\rowmac}c>{\rowmac}c>{\rowmac}c>{\rowmac}c>{\rowmac}c>{\rowmac}c>{\rowmac}c<{\clearrow}}
    \hline
        \multicolumn{5}{c}{\multirow{2}{*}{Cause Effect Estimation}} & \multicolumn{6}{c}{Refutation Tests} \\ \cline{6-11}
        \multicolumn{5}{c}{} & \multicolumn{2}{c}{Placebo} & \multicolumn{2}{c}{RCC} & \multicolumn{2}{c}{RSR} \\ \hline
        Th & Method & ATE & CI & p-value & Effect* & p-value & Effect* & p-value & Effect* & p-value \\\hline
        \setrow{\bfseries} 30 & LR & -0.036 & (-0.059, -0.015) & 0.001 & 0.000 & 0.980 & -0.036 & 0.992 & -0.036 & 0.986 \\
        \rowcolor[HTML]{EFEFEF} \setrow{\bfseries} 30 & M & -0.033 & (-0.056, -0.011) & 0.004 & 0.004 & 0.920 & -0.032 & 0.802 & -0.033 & 0.982 \\
        30 & IPS W & -0.032 & (-0.075, 0.007) & 0.133 & 0.001 & 0.420 & -0.032 & 0.787 & -0.032 & 0.938 \\
        \rowcolor[HTML]{EFEFEF} \setrow{\bfseries} 30 & T-L & -0.050 & (-0.073, -0.024) & 0.001 & 0.000 & 0.480 & -0.047 & 0.009 & -0.047 & 0.614 \\
        \setrow{\bfseries} 30 & X-L & -0.042 & (-0.062, -0.017) & 0.001 & -0.001 & 0.490 & -0.047 & 0.009 & -0.042 & 0.614 \\
        \rowcolor[HTML]{EFEFEF} \setrow{\bfseries} 45 & LR & -0.031 & (-0.053, -0.008) & 0.008 & 0.000 & 0.003 & -0.031 & 0.994 & -0.031 & 0.937 \\
        \setrow{\bfseries} 45 & M & -0.025 & (-0.050, -0.006) & 0.030 & 0.002 & 0.034 & -0.029 & 0.361 & -0.027 & 0.737 \\
        \rowcolor[HTML]{EFEFEF} 45 & IPS W & -0.009 & (-0.047, 0.031) & 0.658 & 0.000 & 0.401 & -0.026 & 0.000 & -0.026 & 0.003 \\
        \setrow{\bfseries} 45 & T-L & -0.039 & (-0.062, -0.014) & 0.002 & 0.000 & 0.456 & -0.038 & 0.501 & -0.039 & 0.918 \\
        \rowcolor[HTML]{EFEFEF} \setrow{\bfseries} 45 & X-L & -0.034 & (-0.062, -0.014) & 0.002 & 0.000 & 0.420 & -0.034 & 0.511 & -0.034 & 0.932 \\
        60 & LR & -0.008 & (-0.032, 0.013) & 0.509 & -0.000 & 0.488 & -0.008 & 0.993 & -0.008 & 0.973 \\
        \rowcolor[HTML]{EFEFEF} 60 & M & -0.004 & (-0.032, 0.014) & 0.731 & -0.000 & 0.744 & -0.010 & 0.177 & -0.007 & 0.698 \\
        60 & IPS W & -0.003 & (-0.040, 0.033) & 0.886 & -0.000 & 0.808 & -0.003 & 0.289 & -0.003 & 0.918 \\
        \rowcolor[HTML]{EFEFEF} 60 & T-L & -0.024 & (-0.043, 0.003) & 0.105 & -0.000 & 0.124 & -0.024 & 0.621 & -0.024 & 0.928 \\
        \setrow{\bfseries} 60 & X-L & -0.030 & (-0.050, -0.012) & 0.002 & -0.000 & 0.124 & -0.030 & 0.621 & -0.030 & 0.928 \\
        \rowcolor[HTML]{EFEFEF} 75 & LR & 0.025 & (-0.028, 0.060) & 0.252 & -0.001 & 0.197 & 0.025 & 0.978 & 0.024 & 0.979 \\
        75 & M & 0.021 & (-0.013, 0.055) & 0.236 & -0.004 & 0.294 & 0.023 & 0.796 & 0.021 & 0.943 \\
        \rowcolor[HTML]{EFEFEF} 75 & IPS W & 0.039 & (-0.018, 0.099) & 0.187 & -0.002 & 0.038 & 0.027 & 0.001 & 0.027 & 0.215 \\
        75 & T-L & -0.001 & (-0.022, 0.019) & 0.120 & -0.001 & 0.126 & -0.001 & 0.217 & -0.001 & 0.839 \\
        \rowcolor[HTML]{EFEFEF} 75 & X-L & 0.001 & (-0.025, 0.023) & 0.957 & -0.001 & 0.126 & 0.002 & 0.216 & 0.002 & 0.839 \\
        \hline
    \end{tabular}
    \label{tab:s_ate_results}
\end{table*}

\clearrow
\begin{table*}[h]
    \centering
    \leftskip-1.5cm
    \caption{{\bf Treatment threshold, method, ATE estimates, 95\% confidence intervals and p-values. Refutation tests fail if their p-value is less than 0.05. IPC-based dataset.}}
    \footnotesize
    \begin{tabular}{>{\rowmac}c>{\rowmac}c>{\rowmac}c>{\rowmac}c>{\rowmac}c>{\rowmac}c>{\rowmac}c>{\rowmac}c>{\rowmac}c>{\rowmac}c>{\rowmac}c<{\clearrow}}
    \hline
        \multicolumn{5}{c}{\multirow{2}{*}{Cause Effect Estimation}} & \multicolumn{6}{c}{Refutation Tests} \\ \cline{6-11}
        \multicolumn{5}{c}{} & \multicolumn{2}{c}{Placebo} & \multicolumn{2}{c}{RCC} & \multicolumn{2}{c}{RSR} \\ \hline
        Th & Method & ATE & CI & p-value & Effect* & p-value & Effect* & p-value & Effect* & p-value \\\hline
        \setrow{\bfseries} 30 & LR & -0.023 & (-0.044, -0.004) & 0.018 & 0.000 & 0.002 & -0.023 & 0.995 & -0.023 & 0.989 \\
        \rowcolor[HTML]{EFEFEF} \setrow{\bfseries} 30 & M & -0.029 & (-0.051, -0.008) & 0.007 & 0.002 & 0.020 & -0.030 & 0.574 & -0.028 & 0.913 \\
        30 & IPS W & -0.015 & (-0.056, 0.027) & 0.470 & 0.000 & 0.201 & -0.021 & 0.001 & -0.021 & 0.003 \\
        \rowcolor[HTML]{EFEFEF} \setrow{\bfseries} 30 & T-L & -0.041 & (-0.066, -0.018) & 0.001 & -0.000 & 0.430 & -0.041 & 0.927 & -0.041 & 0.974 \\
        \setrow{\bfseries} 30 & X-L & -0.039 & (-0.056, -0.020) & 0.001 & -0.001 & 0.460 & -0.039 & 0.927 & -0.039 & 0.974 \\
        \rowcolor[HTML]{EFEFEF} \setrow{\bfseries} 45 & LR & -0.028 & (-0.046, -0.009) & 0.005 & 0.000 & 0.001 & -0.028 & 0.992 & -0.028 & 0.964 \\
        \setrow{\bfseries} 45 & M & -0.026 & (-0.047, -0.008) & 0.006 & 0.001 & 0.017 & -0.026 & 0.479 & -0.024 & 0.829 \\
        \rowcolor[HTML]{EFEFEF} 45 & IPS W & -0.018 & (-0.056, 0.023) & 0.391 & -0.000 & 0.249 & -0.024 & 0.005 & -0.024 & 0.040 \\
        \setrow{\bfseries} 45 & T-L & -0.037 & (-0.057, -0.015) & 0.001 & -0.000 & 0.440 & -0.036 & 0.936 & -0.037 & 0.954 \\
        \rowcolor[HTML]{EFEFEF} \setrow{\bfseries} 45 & X-L & -0.030 & (-0.046, -0.013) & 0.001 & -0.000 & 0.450 & -0.030 & 0.936 & -0.031 & 0.954 \\
        60 & LR & -0.015 & (-0.035, 0.006) & 0.167 & -0.000 & 0.123 & -0.015 & 0.998 & -0.015 & 0.984 \\
        \rowcolor[HTML]{EFEFEF} 60 & M & -0.019 & (-0.041, 0.003) & 0.090 & 0.000 & 0.106 & -0.019 & 0.928 & -0.019 & 0.952 \\
        60 & IPS W & -0.011 & (-0.048, 0.029) & 0.636 & 0.000 & 0.352 & -0.017 & 0.002 & -0.017 & 0.043 \\
        \rowcolor[HTML]{EFEFEF} \setrow{\bfseries} 60 & T-L & -0.021 & (-0.046, -0.006) & 0.015 & -0.000 & 0.057 & -0.021 & 0.702 & -0.021 & 0.979 \\
        \setrow{\bfseries} 60 & X-L & -0.021 & (-0.040, -0.006) & 0.016 & 0.000 & 0.490 & -0.021 & 0.702 & -0.021 & 0.979 \\
        \rowcolor[HTML]{EFEFEF} 75 & LR & -0.004 & (-0.028, 0.020) & 0.717 & 0.000 & 0.607 & -0.004 & 0.996 & -0.004 & 0.983 \\
        75 & M & -0.004 & (-0.032, 0.027) & 0.808 & -0.000 & 0.680 & -0.006 & 0.534 & -0.005 & 0.825 \\
        \rowcolor[HTML]{EFEFEF} 75 & IPS W & 0.006 & (-0.030, 0.042) & 0.758 & 0.000 & 0.726 & 0.001 & 0.099 & 0.000 & 0.522 \\
        75 & T-L & -0.006 & (-0.031, 0.018) & 0.613 & -0.000 & 0.599 & -0.006 & 0.869 & -0.006 & 0.980 \\
        \rowcolor[HTML]{EFEFEF} 75 & X-L & -0.012 & (-0.032, 0.010) & 0.259 & -0.000 & 0.599 & -0.012 & 0.869 & -0.012 & 0.980 \\
        \hline
    \end{tabular}
    \label{tab:ipc_ate_results}
\end{table*}

\clearrow
\begin{table*}[h]
    \centering
    \leftskip-1.5cm
    \caption{{\bf Treatment threshold, method, ATE estimates, 95\% confidence intervals and p-values. Refutation tests fail if their p-value is less than 0.05. Monthly dataset.}}
    \footnotesize
    \begin{tabular}{>{\rowmac}c>{\rowmac}c>{\rowmac}c>{\rowmac}c>{\rowmac}c>{\rowmac}c>{\rowmac}c>{\rowmac}c>{\rowmac}c>{\rowmac}c>{\rowmac}c<{\clearrow}}
    \hline
        \multicolumn{5}{c}{\multirow{2}{*}{Cause Effect Estimation}} & \multicolumn{6}{c}{Refutation Tests} \\ \cline{6-11}
        \multicolumn{5}{c}{} & \multicolumn{2}{c}{Placebo} & \multicolumn{2}{c}{RCC} & \multicolumn{2}{c}{RSR} \\ \hline
        Th & Method & ATE & CI & p-value & Effect* & p-value & Effect* & p-value & Effect* & p-value \\\hline
        \setrow{\bfseries} 30 & LR & -0.024 & (-0.043, -0.003) & 0.024 & 0.000 & 0.980 & -0.025 & 0.996 & -0.024 & 0.988 \\
        \rowcolor[HTML]{EFEFEF} \setrow{\bfseries} 30 & M & -0.026 & (-0.049, -0.006) & 0.012 & 0.001 & 0.980 & -0.029 & 0.465 & -0.026 & 0.876 \\
        30 & IPS W & -0.012 & (-0.050, 0.026) & 0.541 & -0.000 & 0.217 & -0.017 & 0.001 & -0.017 & 0.009 \\
        \rowcolor[HTML]{EFEFEF} \setrow{\bfseries} 30 & T-L & -0.021 & (-0.038, -0.018) & 0.001 & -0.000 & 0.490 & -0.021 & 0.942 & -0.021 & 0.976 \\
        \setrow{\bfseries} 30 & X-L & -0.024 & (-0.037, -0.018) & 0.001 & -0.000 & 0.500 & -0.024 & 0.942 & -0.024 & 0.976 \\
        \rowcolor[HTML]{EFEFEF} \setrow{\bfseries} 45 & LR & -0.026 & (-0.043, -0.009) & 0.006 & 0.000 & 0.940 & -0.026 & 0.996 & -0.026 & 0.966 \\
        \setrow{\bfseries} 45 & M & -0.025 & (-0.046, -0.008) & 0.007 & 0.001 & 0.900 & -0.025 & 0.516 & -0.023 & 0.845 \\
        \rowcolor[HTML]{EFEFEF} 45 & IPS W & -0.016 & (-0.052, 0.019) & 0.376 & -0.000 & 0.268 & -0.022 & 0.004 & -0.022 & 0.046 \\
        \setrow{\bfseries} 45 & T-L & -0.036 & (-0.055, -0.015) & 0.001 & -0.000 & 0.480 & -0.035 & 0.951 & -0.036 & 0.962 \\
        \rowcolor[HTML]{EFEFEF} \setrow{\bfseries} 45 & X-L & -0.033 & (-0.045, -0.024) & 0.001 & -0.000 & 0.480 & -0.035 & 0.951 & -0.036 & 0.962 \\
        60 & LR & -0.013 & (-0.032, 0.006) & 0.187 & 0.000 & 0.139 & -0.013 & 0.999 & -0.013 & 0.984 \\
        \rowcolor[HTML]{EFEFEF} 60 & M & -0.016 & (-0.037, 0.003) & 0.096 & 0.000 & 0.119 & -0.016 & 0.949 & -0.016 & 0.961 \\
        60 & IPS W & -0.009 & (-0.044, 0.026) & 0.620 & 0.000 & 0.337 & -0.015 & 0.003 & -0.014 & 0.044 \\
        \rowcolor[HTML]{EFEFEF} 60 & T-L & -0.019 & (-0.041, 0.001) & 0.068 & -0.000 & 0.380 & -0.019 & 0.440 & -0.019 & 0.280 \\
        \setrow{\bfseries} 60 & X-L & -0.019 & (-0.032, -0.011) & 0.001 & -0.000 & 0.430 & -0.019 & 0.710 & -0.019 & 0.979 \\
        \rowcolor[HTML]{EFEFEF} 75 & LR & -0.006 & (-0.028, 0.016) & 0.580 & 0.000 & 0.629 & -0.006 & 0.996 & -0.006 & 0.984 \\
        75 & M & -0.004 & (-0.030, 0.023) & 0.770 & -0.000 & 0.672 & -0.006 & 0.558 & -0.005 & 0.838 \\
        \rowcolor[HTML]{EFEFEF} 75 & IPS W & 0.006 & (-0.028, 0.041) & 0.729 & 0.000 & 0.729 & 0.001 & 0.119 & 0.000 & 0.556 \\
        75 & T-L & -0.006 & (-0.026, 0.014) & 0.558 & -0.000 & 0.340 & -0.006 & 0.190 & -0.006 & 0.490 \\
        \rowcolor[HTML]{EFEFEF} 75 & X-L & -0.018 & (-0.032, 0.009) & 0.106 & -0.000 & 0.588 & -0.018 & 0.891 & -0.018 & 0.982 \\
        \hline
    \end{tabular}
    \label{tab:m_ate_results}
\end{table*}

\end{appendices}

\end{document}